\documentclass[10pt,twocolumn,letterpaper]{article}

\usepackage[final,algorithms]{wacv}

\usepackage{graphicx}
\usepackage{amsmath}
\usepackage{amssymb}
\usepackage{booktabs}

\usepackage[table]{xcolor}
\definecolor{lightblue}{rgb}{0.4, 0.7, 0.9}
\usepackage{colortbl}  

\usepackage{multirow}
\usepackage{multicol}
\usepackage{tabularx}
\usepackage{siunitx}
\usepackage{algorithm}
\usepackage{algorithmic}
\usepackage{paralist}
\usepackage[pagebackref,breaklinks,colorlinks,bookmarks=false]{hyperref}

\usepackage[capitalize]{cleveref}
\crefname{section}{Sec.}{Secs.}
\Crefname{section}{Section}{Sections}
\Crefname{table}{Table}{Tables}
\crefname{table}{Tab.}{Tabs.}

\def\wacvPaperID{460}
\def\confName{WACV}
\def\confYear{2025}
\begin{document}

\title{Active Learning for Vision-Language Models}

\author{Bardia Safaei\\
Johns Hopkins University\\
{\tt\small bsafaei1@jhu.edu}
\and
Vishal M. Patel\\
Johns Hopkins University\\
{\tt\small vpatel36@jhu.edu}
}
\maketitle

\begin{abstract}
   Pre-trained vision-language models (VLMs) like CLIP have demonstrated impressive zero-shot performance on a wide range of downstream computer vision tasks. However, there still exists a considerable performance gap between these models and a supervised deep model trained on a downstream dataset. To bridge this gap, we propose a novel active learning (AL) framework that enhances the zero-shot classification performance of VLMs by selecting only a few informative samples from the unlabeled data for annotation during training. To achieve this, our approach first calibrates the predicted entropy of VLMs and then utilizes a combination of self-uncertainty and neighbor-aware uncertainty to calculate a reliable uncertainty measure for active sample selection. Our extensive experiments show that the proposed approach outperforms existing AL approaches on several image classification datasets, and significantly enhances the zero-shot performance of VLMs.
\end{abstract}

\section{Introduction}
The rise of foundational vision-language models (VLM) has enabled impressive progress in various tasks in the field of computer vision \cite{radford2021learning, li2022blip, singh2022flava, luddecke2022image, wei2021aligning, yang2023alip}. These models are pretrained on a large collection of image-text pairs and are typically trained using contrastive learning objectives. For example, both CLIP \cite{radford2021learning} and ALIGN \cite{wei2021aligning} formulate their learning objectives as contrastive losses, which learn to bring images and their corresponding textual descriptions closer in the feature space while pushing unmatched pairs apart. By pretraining at a large scale, these models learn a broad understanding of visual concepts enabling them to effectively transfer to numerous downstream tasks.  Furthermore, their pretraining enables generalization and zero-shot learning capabilities that surpass those of models trained under supervision on more limited datasets.

\begin{figure}[t!]
    \begin{center}
        \includegraphics[width=1\linewidth, trim={0cm 0cm 0cm 0cm}]{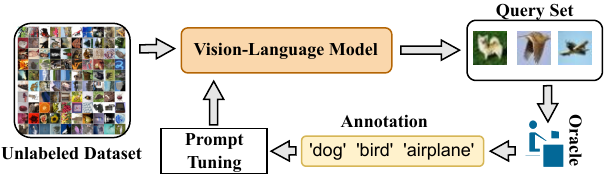}
    \end{center}
    \vskip -20pt
    \caption{
    VL models such as CLIP can effectively transfer to various downstream vision tasks. However, their performance on a novel target dataset still falls behind a supervised model specifically trained on the target dataset. Active learning approaches aim to reduce this performance gap by querying only a few beneficial samples from the unlabeled data, acquiring their labels, and efficiently utilizing them for improving the VLM performance.
    }
    \label{fig:motivation} 
    \vskip -10.0pt
\end{figure}

Although pretrained VLMs have demonstrated their effectiveness in zero-shot settings, their performance on domain-specific datasets is comparatively low when compared to the supervised models specifically trained on those datasets. Therefore, fine-tuning these foundational models on specialized datasets to enhance their performance is a promising direction. Prompt tuning \cite{zhou2022learning, khattak2023maple, zhu2023prompt, jia2022visual} has become an important line of research in recent years for efficiently adapting pre-trained VLMs to new tasks. Context Optimization (CoOp)  \cite{zhou2022learning} is a representative prompt tuning approach that replaces context tokens within a prompt into a set of learnable vectors while keeping the weights of CLIP's image and text encoders frozen. Thus, CoOp requires only a minimal set of labeled images for training and has shown significant improvements compared to manually tuned prompts across various image recognition datasets. 

Despite these advancements, prompt tuning methods typically employ the few-shot setting for fine-tuning the prompts. However, this setting assumes access to a few labeled samples from each class, which may not hold in real-world scenarios. For instance, in a time-sensitive task where a stream of unlabeled data becomes available for immediate training, labeling data until all classes are covered can be costly. Also, the randomly selected few-shot samples may not provide useful and complete information about some datasets. In contrast, active learning (AL) \cite{settles2009active, ren2021survey} aims to select the most informative samples from the unlabeled data without requiring access to samples from each class. As a result, AL is a promising approach for more efficient adaptation of vision-language models by focusing on the most beneficial samples, making it both cost-effective and scalable (see Fig. \ref{fig:motivation}).

However, it has been shown that large pre-trained models like CLIP often produce uncalibrated outputs \cite{levine2023enabling, zhao2021calibrate}, which can lead to imbalanced predictions, with some classes being predicted less frequently than others. This issue negatively impacts conventional uncertainty-based AL methods, as they rely on miscalibrated outputs to calculate uncertainty, resulting in suboptimal performance. 
To address this, we propose \textbf{C}alibrated \textbf{E}ntropy-weighted \textbf{C}lustering (\emph{CEC}), a novel AL framework designed to enhance the selection of informative samples from the unlabeled data which after adding to the labeled training data, can maximally boost the performance of prompt tuning approaches in VLMs. Our method utilizes a calibrated uncertainty score, followed by a novel clustering approach that enables sampling from more vulnerable regions in the feature space. Specifically, we calibrate the predictive entropy of the CLIP model to reduce the bias towards the more frequently encountered categories. This entropy score quantifies the uncertainty for a single sample. Additionally, we leverage CLIP's rich representations by proposing a neighbor uncertainty measure for each sample that enhances the reliability of uncertainty estimates. Finally, to ensure diversity, we adaptively cluster and perform an uncertainty-weighted sampling on each cluster according to the AL budget. Our experiments show that our method outperforms existing state-of-the-art AL methods on several image classification datasets and significantly improves the zero-shot performance of VLMs. 

The contributions of this paper can be summarized as follows.
\begin{itemize}
    \item In this paper, we systematically analyze and show the advantages of utilizing AL strategies for prompt tuning of VLMs. In particular, we propose an AL method for VLMs that selects beneficial samples for prompt tuning and significantly improves the zero-shot performance of VLMs.
    \item We introduce an AL framework that leverages both self-uncertainty and neighbor-aware uncertainty of unlabeled samples to select informative samples during AL rounds.
     
    \item Our experimental results show that the proposed method outperforms existing state-of-the-art methods on several image classification datasets.
\end{itemize}

\section{Related Work}
\noindent {\bf{Vision-Language Models (VLM).}} Vision-Language Pre-training (VLP) has emerged as a promising method for developing transferable and versatile recognition models by establishing connections between visual content and language descriptors. This approach has been explored in various studies \cite{joulin2016learning, li2021supervision,gan2022vision,chen2023vlp, yang2023alip}, where the primary challenge is the limited size of the training datasets, such as Flickr \cite{joulin2016bag} and COCO Captions \cite{desai2021virtex}. However, recent advancements in VLP, exemplified by models like CLIP \cite{radford2021learning} and ALIGN \cite{jia2021scaling} models, have achieved impressive outcomes. These models utilize web-scale noisy image-text pairs and employ a contrastive objective that aligns matching image-text pairs while distancing non-matching ones. Unlike ALIGN and CLIP, which mostly focus on specific visual tasks, FLAVA \cite{singh2022flava} is a universal VLM targeting vision, language, and multi-modal tasks. BLIP \cite{li2022blip} is a multi-modal mixture of encoder-decoder, which is pretrained on a bootstrapped dataset to obtain improved performance. BLIP-2 \cite{li2023blip} enhances the cost-efficiency of BLIP by keeping image and language encoders frozen and pretraining a lightweight transformer to reduce the modality gap. SLIP \cite{mu2022slip} incorporates self-supervision into the multi-modal pretraining objective. By harnessing the power of natural language supervision, these VLMs not only obtain robust visual representations but also show remarkable adaptability to a wide range of downstream applications.\\
\noindent {\bf{Prompt Tuning (PT).}} Recent advancements in natural language processing (NLP) have given rise to a novel paradigm known as prompt learning/engineering \cite{gao2020making, brown2020language, shin2020autoprompt, zhang2021vinvl,li2021prefix,ye2022ontology}, which aims to optimize learnable prompts instead of end-to-end fine-tuning of the model. has gradually supplanted the traditional fine-tuning approach in NLP. Recently, there have been enormous research efforts to develop prompt learning approaches for fine-tuning VLMs \cite{zhou2022learning, jia2022visual, khattak2023maple, lu2022prompt, zhang2023adversarial}. For example, CoOp \cite{zhou2022learning} replaces context tokens within a prompt into a set of learnable vectors while keeping the weights of CLIP's image and text encoders frozen. VPT \cite{jia2022visual} learns prompts on the visual side. It introduces learnable parameters to the input sequence of every Transformer layer, and these learnable parameters are jointly optimized with the linear classification layer during fine-tuning. Unlike CoOp and VPT, MaPLe learns the prompts in a multi-modal fashion, utilizing both the vision and language branches. ProDA \cite{lu2022prompt} is another PT method that attempts to learn the category-wise distribution of prompts and utilize those distributions to generate a diverse set of prompts.\\
\begin{figure*}[t!]
    \begin{center}
        \includegraphics[width=0.75\linewidth, trim={0.5cm 0.5cm 0.9cm 0cm}]{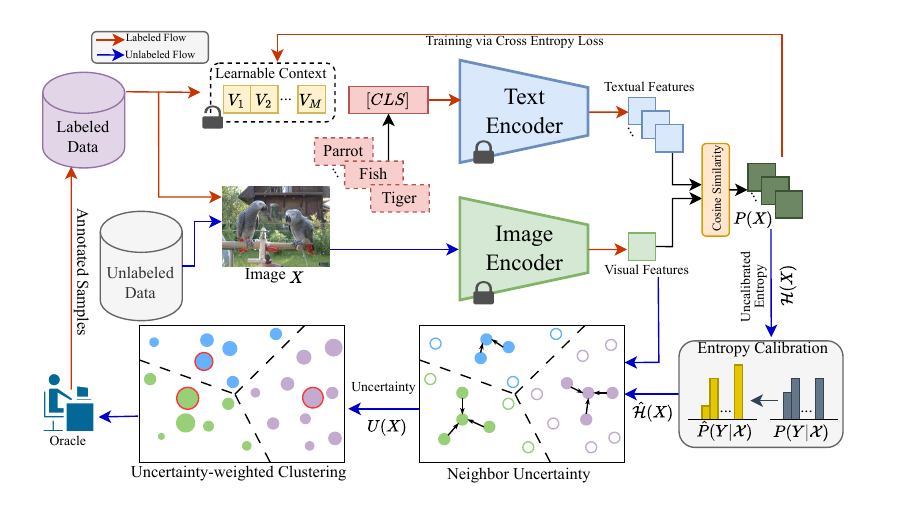}
    \end{center}
    \vskip -20.0pt
    \caption{\textbf{An illustration of our \emph{CEC} framework.} For a given unlabeled dataset, the visual features are extracted, and the prediction probabilities are calculated using textual features. Next, we calibrate the entropy score and utilize it to precisely quantify the uncertainty of an unlabeled sample, considering its similarities to textual embeddings. Moreover, we regularize this entropy by incorporating the uncertainty of a sample's neighbors. Ultimately, this calibrated entropy is integrated into a uncertainty-weighted clustering approach to ensure diverse sample selection. The selected samples are then annotated and used for prompt tuning of CLIP.
    }
    \label{fig:framework} 
\end{figure*}

\noindent {\bf{Active Learning (AL).}} Active learning focuses on enhancing the performance of a deep model on an unlabeled dataset by selecting a few informative samples for supervised training. AL algorithms are particularly important when labeling the entire unlabeled data is impractical and the annotation budget is limited. AL methods can be divided into three main groups, namely uncertainty-based \cite{parvaneh2022active, balcan2007margin, ash2019deep}, diversity-based \cite{sener2017active, ash2019deep, agarwal2020contextual}, and ensemble-based \cite{freund1997selective} approaches. Uncertainty-based methods are based on selecting the most confusing samples for the model by relying on various uncertainty scoring functions, such as entropy \cite{wehrl1978general, safaei2024entropic}, softmax margin \cite{balcan2007margin}, and mutual information \cite{kraskov2004estimating}. Diversity-based approaches \cite{sener2017active, yang2015multi} attempt to choose samples from diverse regions of feature space to better adapt to the distribution of unlabeled data. Ensemble-based methods \cite{freund1997selective, hino2023active} use a measure of disagreement among an ensemble of models with different characteristics (e.g. model initialization) as the active sampling criterion. Lately, there have been methods that improve AL performance by using a hybrid of multiple sample selection criteria \cite{ash2019deep, wei2015submodularity, parvaneh2022active, xie2023towards, prabhu2021active, xie2023active}. For example in  \cite{ash2019deep}, the authors combine diversity and uncertainty for improved performance by clustering gradient representations of unlabeled samples. In \cite{xie2023active}, the authors optimize a parametric model to find the most informative samples for a pretraining-finetuning task. \cite{parvaneh2022active} utilizes a mixup technique to generate novel features for unlabeled data. Samples with confusing novel features are then clustered, and queries are made from different clusters. 
\section{Background}
In this section, we first introduce the problem of AL for prompt tuning of VLMs. Following that, we provide a brief overview of the training process for the CLIP model, followed by an explanation of the prompt tuning approach we employed for fine-tuning the CLIP model.\\
\noindent {\bf{Problem Formulation.}}
In this paper, we consider active learning for a $K$-way image classification task using the CLIP model, where $K$ is the number of classes. In this setting, we begin with a pool of unlabeled data $\mathcal{D}_{U}=\left\{ x_{i}^{U} \right\}_{i=1}^{N_{U}}$ and an initially empty labeled dataset, denoted as  $\mathcal{D}_{L}=\emptyset$. 
At each AL round, the query function selects a batch of $b$ samples $X^{q}$ from the unlabeled data, and their labels are queried from an oracle. We then add these annotated samples to the labeled dataset as $\mathcal{D}_L = \mathcal{D}_L \cup X^q$. The updated $\mathcal{D}_L$ is employed to finetune the CLIP model by tuning CoOp's learnable prompts.

\subsection{Contrastive Language-Image Pretraining}
In this section, we briefly review the well-known Contrastive Language-Image Pretraining (CLIP) model. The CLIP's architecture consists of an image encoder and a text encoder. CLIP utilizes a contrastive loss \cite{chen2020simple} to align the feature vectors generated by these encoders. In particular, for a given batch of images and their corresponding text descriptions, CLIP initially extracts the visual and textual embeddings via encoders. Next, these embeddings undergo normalization, and the model maximizes the cosine similarity between a textual embedding and its corresponding visual embedding. Simultaneously, CLIP minimizes the cosine similarities between unmatched textual and visual embeddings. This approach enables CLIP to learn multimodal information from massive datasets of image-text pairs, contributing to its impressive zero-shot classification performance. During inference, for a given image $x$ belonging to a dataset with $K$ different classes and class names $\{ \textup{CLASS}_i\}_{i=1}^K$, $K$ class-wise prompts are generated as `a photo of a $\{ \textup{CLASS}_i\}$'. By passing these prompts to the text encoder, we obtain class-wise textual embeddings $t_i$. The probability of $x$ belonging to the class $i$ is then calculated as follows:
\setlength{\belowdisplayskip}{0pt} \setlength{\belowdisplayshortskip}{0pt}
\setlength{\abovedisplayskip}{0pt} \setlength{\abovedisplayshortskip}{0pt}
\begin{equation} 
 P(y=i|x) = \frac{\operatorname{exp}(\operatorname{sim}(f(x),t_i)/T)}{\sum_{j=1}^{K} \operatorname{exp}(\operatorname{sim}(f(x),t_j)/T)},
 \label{prob_eq}
\end{equation}
where $f(\cdot)$ denotes the image encoder, $\operatorname{sim}(\cdot, \cdot)$ is the cosine similarity, and $T$ is the temperature scaling value. The class with the highest probability is considered the prediction of $x$. The value of $T$ is set to $0.01$ in the original CLIP paper, and we follow the same setting in this paper.
\subsection{Prompt Tuning for Training}
One of the challenges associated with large multimodal pretrained models, such as CLIP, is fine-tuning them on new training samples while preserving their intrinsic multimodal representations. Typically, two common approaches are considered: end-to-end fine-tuning of the CLIP encoders and linear probing. The end-to-end fine-tuning approach updates all parameters of the model, whereas the linear probing method trains a linear classification layer on top of CLIP's backbone. However, these approaches not only risk diminishing the CLIP's adaptability to new domains but also show instability when employed in scenarios with limited training data. 

To mitigate such limitations, we utilize a prompt tuning approach called Context Optimization (CoOp) \cite{zhou2022learning} in this paper. CoOp replaces the fixed context tokens in prompts with learnable vectors. 

In particular, it uses $M$ learnable context vectors 
$\{V_1, V_2, \hdots, V_M\}$ along with the class token $C_i$ for the $i$-th class name as the prompt. Thus, the textual embedding for class $i$ is calculated as $t_i = h(\left [ V_1,V_2, \hdots, V_M, C_i \right ])$ where $h$ is the text encoder\footnote{\cite{zhou2022learning} has explored other variations such as positioning the class token in the middle of context tokens or using class-specific context tokens. In our experiments, we use the default setup as mentioned.}. CoOp trains learnable context vectors via cross-entropy loss while the parameters of the CLIP backbone are kept frozen. 
\begin{algorithm}[tb]
	\caption{Our proposed \emph{CEC} algorithm for AL}
	\label{alg:LfOSA}
	\begin{algorithmic}[1]
		\STATE \textbf{Input:}
		\STATE \quad Unlabeled data $\mathcal{D}_{U}$, number of AL 
  
  \quad cycles $R$, known categories $K$, per-cycle budget $b$, 
  
  \quad image encoder $f$, text encoder $h$, 
		\STATE \textbf{Process:}
		
		\STATE \quad \textbf{for} $c=0,1,...,R-1$ \textbf{do}
		  \STATE \quad \quad $\forall x \in D_{U}$, $U(x)\gets 0$ \quad \quad \emph{{ \textcolor{lightblue}{\# Initialization}}}
		\STATE \quad \quad Calibrate the predicted entropy of $x$ via Eq. \ref{eq:calibration}.
	\STATE \quad \quad Compute neighbor uncertainty. $\mathcal{H}_{\text{NN}}$ via \ref{eq:prop_final}.
        \STATE \quad \quad Compute $U(x)$ via \ref{eq:final_unc}.
        \STATE \quad \quad \emph{\textcolor{lightblue}{\# Clustering}}
        \STATE \quad \quad Perform Weighted K-Means clustering on features.         
        \STATE \quad \quad For the $i$-th cluster, find the sample $x_i$ closest to 
        \STATE \quad \quad the cluster center. 
        \STATE \quad \quad \emph{{\textcolor{lightblue}{\# All queries for the current cycle:}}}
        \STATE \quad \quad $X^{q} \gets x_1 \cup x_2 \cup ... \cup x_b$ 
        \STATE \quad \quad $\mathcal{D}_{L} \gets \mathcal{D}_{L} \cup X^q$,  
        \STATE \quad \quad $\mathcal{D}_{U} \gets \mathcal{D}_{U} / X^{q}$ \quad \emph{\textcolor{lightblue}{{\# Update datasets}}}
        \STATE \quad \quad \emph{{\textcolor{lightblue}{\# Prompt Tuning}}}
         \STATE \quad \quad Update CoOp's learnable context tokens $V_i$  
         \STATE \quad \quad via cross-entropy on $D_L$
            \STATE \quad \textbf{Return} Updated prompts
            
	\end{algorithmic}
 \label{alg:algorithm}
 
\end{algorithm}

\section{Methodology}
In this section, we provide a detailed elaboration on our proposed approach, Calibrated Entropy-weighted Clustering (\emph{CEC}), which is tailored to the problem of AL for prompt tuning of pre-trained VLMs, such as the CLIP model. In \emph{CEC}, we first utilize a calibrated entropy score to effectively quantify the uncertainty of an unlabeled sample based on its similarities to the textual embeddings. Furthermore, we regularize this entropy by considering the uncertainty of a sample's neighbors to mitigate the problem of outlier selection. Finally, we incorporate our calibrated entropy into a clustering approach to ensure diverse sample selection (see Algorithm \ref{alg:algorithm} and Fig. \ref{fig:framework} for more details).\\
\noindent \textbf{Notation.} We consider a CLIP model with an image encoder $f(\cdot)$ and a text encoder $h(\cdot)$ for a $K$-way classification problem. We denote the textual embedding corresponding to the $i$-th class as $t_i$. 
\vspace{-1.5mm} 
\subsection{Entropy Calibration}
\vspace{-1mm} 
To measure the uncertainty of a given unlabeled image $x$, we utilize CLIP's predicted entropy as follows:
\begin{equation}
\begin{aligned}
        \mathcal{H}(x)=-\sum_{i=1}^{K} P(y=i|x) \cdot \log (P(y=i|x)),
\end{aligned}
\label{ent}
\end{equation}
where $P(y=i|x)$ was defined in Eq. \ref{prob_eq}. However, this entropy can be unreliable given the miscalibration issue inherent in large pretrained models including CLIP \cite{shao2023investigating, yu2022cold, zhao2021calibrate}. To mitigate this issue, we estimate the \textit{contextualized prior} \cite{hu2021knowledgeable} for each class to calibrate the predicted probabilities. For the $i$-th class, we select the first $N$ samples with the highest $P(y=i|x)$ values as follows:
\begin{equation}
    \operatorname{S}_{i} = \left\{  x'|P(y=i|x')\in \operatorname{Top}_{N}(P(y=i|x))\right\},
\end{equation}
where $x,x'\in \mathcal{D}_U$. The contextualized prior for class $i$ is calculated by
\begin{equation}
    \setlength{\abovedisplayskip}{0.2pt}
    \setlength{\belowdisplayskip}{0.2pt}
    Q(i) \approx \frac{1}{N} \sum_{x \in \operatorname{S_i}} P(y=i|x).
\end{equation}
We then calibrate the probabilities as
\begin{equation}
\setlength{\abovedisplayskip}{0.2pt}
\setlength{\belowdisplayskip}{0.2pt}
\hat{P}(y=i|x) = \left({\frac{P(y=i|x)}{Q(i)}}\right) /  \left({\sum_{j=1}^{K} \frac{P(y=j|x)}{Q(j)}}\right). 
\label{eq:calibration}
\end{equation}
Finally, the calibrated entropy $\hat{\mathcal{H}}$ is calculated by substituting Eq. \ref{eq:calibration} to Eq. \ref{ent}.
\vspace{-2mm}
\subsection{Neighbor Uncertainty}
While we have calibrated the entropy of each unlabeled sample, relying only upon $\hat{\mathcal{H}}$ might lead to selecting highly uncertain but outlier samples. To address this problem, for a given sample, we take the uncertainty of its nearest neighbors into consideration. Thus, an unlabeled sample is recognized as uncertain if it has both high self-uncertainty and neighbor uncertainty. Labeling such samples not only resolves their self-uncertainties but also helps reduce the model's confusion in a local neighborhood around
those samples. Formally, we denote the $k$-nearest neighbors of a given sample $x$ in feature space as $k$NN$(x)$.  The neighbor uncertainty is defined as:
\begin{equation}
\setlength{\abovedisplayskip}{0.2pt}
\setlength{\belowdisplayskip}{0.2pt}
\mathcal{H}_{\text{NN}}(x) = \frac{1}{k}\cdot\sum_{x_i\in k\text{NN}(x)}\exp \left(-\alpha\left\|\mathbf{z}-\mathbf{z}_{i}\right\|_2^{2}\right)\cdot \hat{\mathcal{H}}(x_i),
\label{eq:prop_final}
\end{equation}
where $\mathbf{z} = \frac{f(x)}{\left\|f(x)\right\|}$ is the normalized visual representation of $x$. We formulate the uncertainty of a sample $x$ as 
\begin{equation}
\setlength{\abovedisplayskip}{0.2pt}
\setlength{\belowdisplayskip}{0.2pt}
U(x) = \hat{\mathcal{H}}(x) + \mathcal{H}_{\text{NN}}(x).
\label{eq:final_unc}
\end{equation}

\subsection{Uncertainty-weighted Clustering}
Similar to several existing AL methods \cite{parvaneh2022active, ash2019deep, prabhu2021active}, we propose to perform clustering on the visual features to ensure sampling from diverse regions of the feature space. Specifically, we assign weights to visual features based on their corresponding uncertainty (Eq. \ref{eq:final_unc}). Intuitively, this weighting mechanism increases the number of instances from a particular sample proportional to its uncertainty, enhancing the likelihood of its selection from the cluster. Our clustering approach strikes a balance between uncertainty and diversity resulting in significant performance improvement. We have empirically demonstrated that uncertainty-weighted clustering achieves superior performance compared to the alternative approach that clusters the features and selects the most uncertain samples within each cluster (see Fig. \ref{fig:ablation} (right)). Our clustering approach can be implemented using a weighted K-Means \cite{huang2005automated} algorithm.

\subsection{Query Strategy and Training}
As described in previous sections, at each AL cycle, we first calculate the uncertainty of samples via Eq. \ref{eq:final_unc} and then perform uncertainty-weighted clustering to partition samples into $b$ clusters. We query the label for the closest sample to each cluster center. After updating $\mathcal{D}_L$, we finetune CLIP on the labeled samples using the CoOp approach.

\begin{table*}[t!]
\vskip-10pt\caption{\textbf{Results for CoOp \cite{zhou2022learning} Prompt Tuning.} Classification accuracy comparisons on 6 specialized datasets. $B$ is the per-cycle annotation budget. All experiments are conducted across three seeds, and the average results are reported. The second-best results are underlined.}
\label{tab:coop_results}
\renewcommand{\arraystretch}{1}
    \centering
    \resizebox{0.9\textwidth}{!}{$
    \setlength{\tabcolsep}{1mm}{
\begin{tabular}{c|c|cccccc|>{\columncolor{gray!20}}c|c}
\toprule
\textbf{Dataset} & $B (\%)$ &  \textbf{Random} & \textbf{Entropy}\cite{wang2014new} & \textbf{CoreSet} \cite{sener2017active} & \textbf{BADGE} \cite{ash2019deep} & \textbf{ALFA-Mix} \cite{parvaneh2022active} & \textbf{GCNAL} \cite{ash2019deep} & \textbf{Ours} & \textbf{Zero-shot} \\ \hline

\multirow{3}{*}{Textures} & 1  & 38.4 $\pm$ 0.2     & 35.2 $\pm$ 0.8 & - & \underline{40.2 $\pm$ 5.0} & - & 38.8 $\pm$ 0.9 & \textbf{47.9 $\pm$ 1.2} & \multirow{3}{*}{44.3} \\  
                &  2  &  44.2 $\pm$ 2.9    & 40.9 $\pm$ 2.0 & 44.9 $\pm$ 0.9 & 46.9 $\pm$ 1.4 & \underline{49.6 $\pm$ 0.4} & 44.8 $\pm$ 0.4 & \textbf{52.8 $\pm$ 1.0} &    \\  
                &  5  &  54.1 $\pm$ 2.9   & 49.3 $\pm$ 1.4 & 52.9 $\pm$ 2.3 & \underline{56.2 $\pm$ 1.1} & 55.6 $\pm$ 1.9 & 55.0 $\pm$ 1.4 & \textbf{58.2 $\pm$ 2.0} &   \\ \hline  

\multirow{3}{*}{Caltech101} & 1  &  88.2 $\pm$ 3.4   & 86.1 $\pm$ 4.6 & - & 88.2 $\pm$ 1.7 & - & \underline{88.4 $\pm$ 3.3} & \textbf{88.7$\pm$ 1.5} & \multirow{3}{*}{91.3} \\  
                &  2  &  88.4 $\pm$ 3.2    & 89.3 $\pm$ 1.4 & \textbf{91.1 $\pm$ 2.0} & 89.8 $\pm$ 2.7 & 89.7 $\pm$ 0.8 & \underline{89.8 $\pm$ 2.0} & 89.0 $\pm$ 1.3 &         \\  
                &  5  &  91.1 $\pm$ 1.1  & 89.4 $\pm$ 0.8 & 91.3 $\pm$ 0.3 & 92.2 $\pm$ 0.1 & 92.3 $\pm$ 0.3 & \underline{92.4 $\pm$ 0.9} & \textbf{92.8 $\pm$ 0.5} &         \\ \hline 

\multirow{3}{*}{EuroSAT} & 1  &  \underline{82.2 $\pm$ 1.0}  & 70.5 $\pm$ 2.0 & - & 80.6 $\pm$ 0.7 & - & 82.1 $\pm$ 1.4 & \textbf{82.8 $\pm$ 1.6}& \multirow{3}{*}{42.0} \\  
                &  2  &  86.1 $\pm$ 1.0   & 78.1 $\pm$ 4.3 & 84.5 $\pm$ 1.5 & 85.6 $\pm$ 0.7 & \underline{86.1 $\pm$ 0.3} & 84.0 $\pm$ 0.9 & \textbf{86.2 $\pm$ 0.6} &         \\  
                &  5  &  87.8 $\pm$ 0.6   & 84.8 $\pm$ 2.1 & 87.9 $\pm$ 1.4 & 87.5 $\pm$ 0.5 & \textbf{88.3 $\pm$ 0.3} & \underline{88.2$\pm$ 0.6} & 88.0 $\pm$ 0.9 &         \\ \hline 

\multirow{3}{*}{FGVC-Aircraft} & 1  &  18.4 $\pm$ 0.6   & \underline{19.7 $\pm$ 1.1} & - & 17.8 $\pm$ 1.7 & - & 18.4 $\pm$ 0.6 & \textbf{20.3 $\pm$ 1.1}& \multirow{3}{*}{24.9} \\  
                &  2  &  21.2 $\pm$ 1.4    & \underline{22.0 $\pm$ 2.1} & 19.7 $\pm$ 1.4 & 20.7 $\pm$ 1.1 & 20.2 $\pm$ 1.5 & 21.2 $\pm$ 0.4 & \textbf{22.3 $\pm$ 1.0} &         \\  
                &  5  &  26.0 $\pm$ 1.0   & 24.7 $\pm$ 0.6 & 23.0 $\pm$ 0.1 & 25.7 $\pm$ 1.0 & \textbf{28.7 $\pm$ 0.4} & 26.3 $\pm$ 0.7 & \underline{27.1 $\pm$ 0.3} &         \\ \hline 

\multirow{3}{*}{Flowers102} & 1  &  \underline{60.2 $\pm$ 2.2}    & 55.2 $\pm$ 4.7 & - & 53.5 $\pm$ 5.3 & - & 60.2 $\pm$ 2.3 & \textbf{64.1 $\pm$ 2.4} & \multirow{3}{*}{67.3} \\  
                &  2  &  66.3 $\pm$ 2.2    & 65.4 $\pm$ 3.5 & 62.2 $\pm$ 0.7 & 68.2 $\pm$ 1.7 & \underline{74.0 $\pm$ 0.8} & 64.0 $\pm$ 4.0 & \textbf{75.6 $\pm$ 2.5} &         \\  
                &  5  &  82.6 $\pm$ 2.2   & 80.3 $\pm$ 2.8 & 76.2 $\pm$ 2.6 & 86.2 $\pm$ 1.6 & \textbf{88.7 $\pm$ 1.2} & 80.0 $\pm$ 2.9 & \underline{88.2 $\pm$ 1.6} &         \\ \hline 

\multirow{3}{*}{UCF101} & 1  &  \underline{55.4 $\pm$ 2.7}   & 53.1 $\pm$ 3.9 & - & 50.7 $\pm$ 3.0 & - &55.3 $\pm$ 3.7 & \textbf{57.6 $\pm$ 1.8}& \multirow{3}{*}{64.3} \\  
                &  2  &  66.6 $\pm$ 1.2   & 62.1 $\pm$ 1.9 & 65.4 $\pm$ 1.4 & 63.7 $\pm$ 1.5 & \underline{66.9 $\pm$ 1.5} & 63.9 $\pm$1.9 & \textbf{67.0 $\pm$ 0.8} &         \\  
                &  5  &  73.8 $\pm$ 0.4   & 72.8 $\pm$ 0.7 & 74.1 $\pm$ 3.0 & 75.3 $\pm$ 1.2 & \underline{75.4 $\pm$ 1.9} & 73.1$\pm$ 1.4 & \textbf{76.2 $\pm$ 0.6}&         \\ \hline

\multirow{3}{*}{\textit{Average}} & 1  &  57.1   & 53.3 & - & 55.2 & - & \underline{57.2} & \textbf{60.2} (\textcolor{blue}{+3.0}) & \multirow{3}{*}{55.7} \\  
                &  2  &  62.1   & 59.6 & 61.3 & 62.5 & \underline{64.4} & 61.3 & \textbf{65.5} (\textcolor{blue}{+1.1})&         \\  
                &  5  &  69.2   & 66.9 & 67.6 & 70.5 & \underline{71.5} & 69.2 & \textbf{71.8} (\textcolor{blue}{+0.3})&         \\
\bottomrule
\end{tabular}}
$}
\label{tab:results}
\end{table*}
 
\section{Experimental Setup}
\noindent {\bf{Datasets.}} Following prompt tuning works \cite{khattak2023maple,zhou2022conditional}, we select 6 different image classification datasets for our experiments, namely Describable Textures \cite{cimpoi14describing}, Caltech-101 \cite{bansal2021transfer}, EuroSAT \cite{helber2019eurosat}, FGVC-Aircraft \cite{maji2013fine}, Flowers-102 \cite{kanan2010robust}, and UCF-101 \cite{soomro2012ucf101}. These datasets cover a range of specialized computer vision tasks that are suitable for evaluating a large pre-trained model like CLIP that has zero-shot capabilities. More information about these datasets can be found in the supplementary material.\\
\noindent {\bf{Implementation Details.}} In all experiments, we use a pretrained vision transformer ViT-B16 \cite{dosovitskiy2010image} model as the CLIP's backbone. For CoOp prompt tuning, we initialize prompts with `a photo of a $\left \{  \right \}$'. Also, we do not use class-specific context tokens, and the position of the class token is set to `end'. In each AL round, we load the zero-shot CLIP model and fine-tune it on the labeled data. We train for 200 epochs using the SGD optimizer \cite{keskar2017improving} with an initial learning rate of 0.002, a momentum of 0.9, a weight decay of 0.005, and a cosine annealing scheduler. The batch size is set to 32 for all experiments.  We fixed the value of N to 10. We utilize an NVIDIA A5000 GPU to run each experiment.

\section{Experiments}
We compare our method against a suite of state-of-the-art AL approaches, namely, \textbf{ALFA-Mix \cite{parvaneh2022active}}, \textbf{BADGE \cite{ash2019deep}}, \textbf{GCNAL \cite{caramalau2021sequential}}, \textbf{CoreSet \cite{sener2017active}}, \textbf{Entropy \cite{wang2014new}}, and \textbf{Random}. More details about these methods can be found in the supplementary material. 

In our experiments, each method is applied within the same experimental setup to ensure a fair comparison. All datasets are split into train, validation, and test sets, and the train set is used as the unlabeled dataset. The effectiveness of each approach is measured by the model's performance on the held-out test set after each round of AL. 

\noindent
\textbf{Active Learning Setting.}
In all experiments, we perform 6 rounds of AL, where at each round we select $1\%$ of the unlabeled data for annotation. For AL methods that require initial labeled data we perform random sampling in the first round. For a fair comparison, we conduct each experiment across three different seeds and report the mean and standard deviations of the results.   

\subsection{Main Results}
\begin{figure*}[t!]
    \begin{center}
        \includegraphics[width=0.85\linewidth, trim={0cm 0cm 1cm 0cm}]{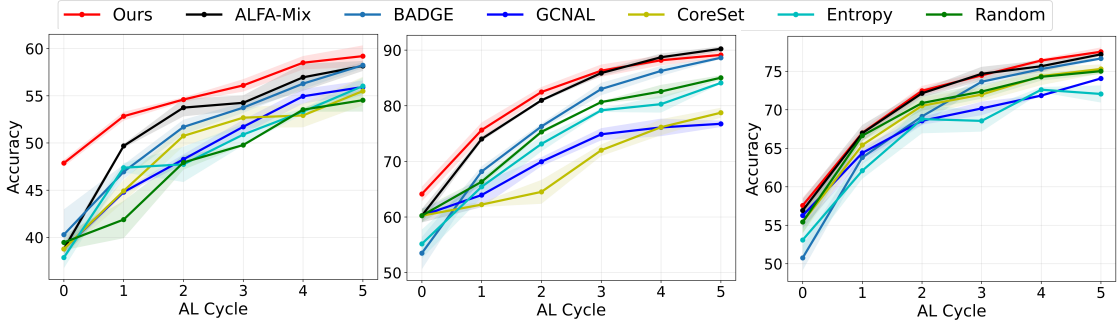}
    \end{center}
    \vskip -12.0pt
    \caption{ CoOp accuracy results over different 6 AL cycles. From left to right: Textures, Flowers102, and UCF101 datasets.}
    \label{fig:result_fig} 
    \vskip -10.0pt
\end{figure*}

\noindent \textbf{CoOp results.} Table \ref{tab:coop_results} and Fig. \ref{fig:result_fig} show the classification results corresponding to the CoOp prompt tuning method on six image classification datasets. In the table, we compare our method with uncertainty-based, diversity-based, hybrid, random, and zero-shot approaches. 
\begin{figure}[t!]
    \begin{center}
        \includegraphics[width=0.80\linewidth, trim={3cm 1cm 2cm 0cm}]{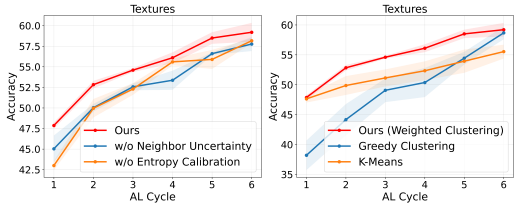}
    \end{center}
    \vskip -15.0pt
    \caption{ Effect of each component within our framework.
    }
    \label{fig:ablation} 
    \vskip -5.0pt
\end{figure}

\begin{figure}[t!]
\vspace{-3mm}
    \begin{center}
        \includegraphics[width=0.80\linewidth]{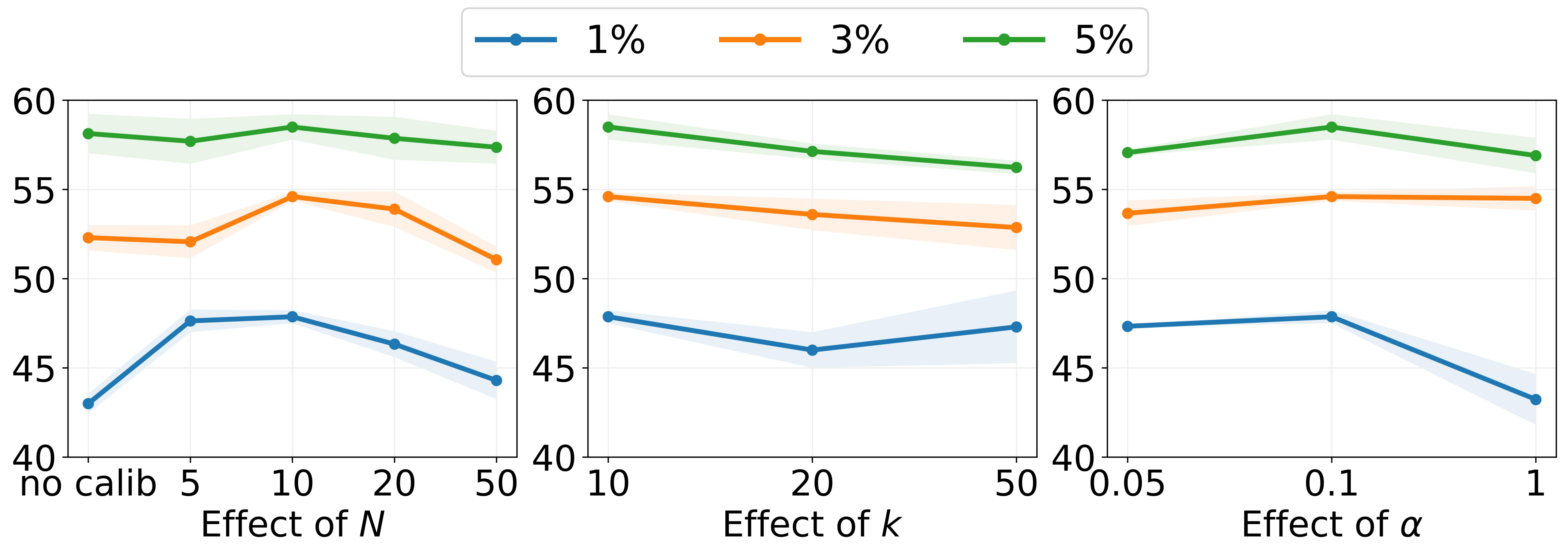}
    \end{center}
    \vspace{-7mm}
    \caption{ Hyperparameter analysis.
    }
    \label{fig:hyper_param} 
    \vspace{-5mm}
\end{figure}
\begin{figure*}[t!]
    \begin{center}
        \includegraphics[width=0.85\linewidth, trim={3cm 1cm 2cm 0cm}]{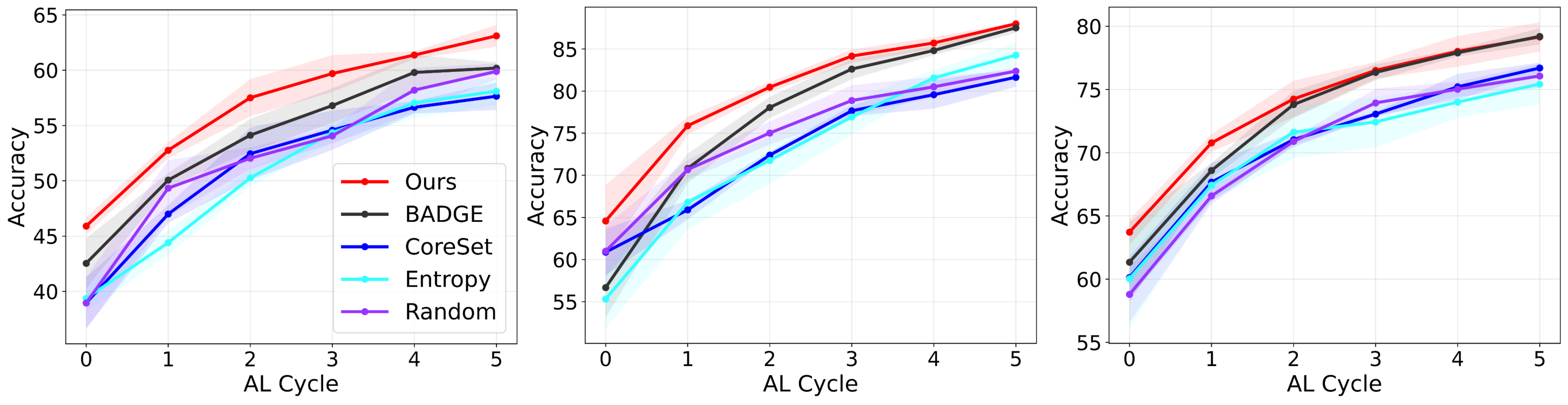}
    \end{center}
    \vskip -12.0pt
    \caption{VPT accuracy results over 6 AL cycles. From left to right: Textures, Flowers102, and UCF101 datasets.}
    \label{fig:vpt_curve} 
    \vskip -10.0pt
\end{figure*}

The results are presented for $1\%$, $2\%$, and $5\%$ annotation budgets. It can be seen that the proposed approach outperforms other baselines in most of the annotation ratios and datasets. In particular, on the Textures dataset, our method outperforms the SOTA method ALFA-Mix by margins of $3.2\%$ and $2.6\%$ for budgets $2\%$ and $5\%$, respectively. Furthermore, it outperforms random sampling by large margins of $9.5\%$ $8.6\%$ and $4.1\%$ for different annotation budgets on the Textures dataset. It can be observed that uncertainty-based approaches such as Entropy and GCNAL underperform random sampling in some budgets. This can be due to uncalibrated uncertainty utilization. However, our method consistently outperforms random sampling which shows the effectiveness of our approach. Moreover, Fig. \ref{fig:result_fig} shows the performance of AL approaches over different AL cycles for Textures, Flowers102, and UCF101 datasets. Our method maintains a superior performance compared to other baselines in almost all rounds.\\ 
\noindent \textbf{Results for other prompt tuning methods.}
To verify the generalization ability of our proposed method, we show the classification results for two additional prompt tuning algorithms, namely VPT \cite{jia2022visual} and MaPLe \cite{khattak2023maple}. From the results in Table \ref{tab:combined_results} and Fig. \ref{fig:vpt_curve}, it is evident that our approach achieves superior performance regardless of the prompt tuning approach. Furthermore, the proposed method significantly improves the zero-shot performance of CLIP. For example in the MaPLe setting, our approach outperforms zero-shot CLIP by $0.6\%$, $7.7\%$, and $15\%$ when sampling $1\%$, $2\%$, and $5\%$ of the unlabeled samples, respectively.
\begin{table*}[t!]
\vskip-10pt\caption{\textbf{Results for VPT \cite{jia2022visual} and MaPLe \cite{khattak2023maple} Prompt Tuning Methods.}}
\label{tab:combined_results}
\renewcommand{\arraystretch}{1}
    \centering
    \resizebox{0.7\textwidth}{!}{$
    \setlength{\tabcolsep}{1.2mm}{
\begin{tabular}{c|c|c|cccc|>{\columncolor{gray!20}}c|c}
\toprule
\textbf{Method} & \textbf{Dataset} & $B (\%)$ &  \textbf{Random} & \textbf{Entropy} \cite{wang2014new} & \textbf{CoreSet} \cite{sener2017active} & \textbf{BADGE} \cite{ash2019deep} &  \textbf{Ours} & \textbf{Zero-shot} \\ \hline

\multirow{12}{*}{\centering\rotatebox{90}{VPT}} & \multirow{3}{*}{Textures}  & 1  & 38.9 $\pm$ 2.9     & 39.3 $\pm$ 1.5 & - & \underline{42.5 $\pm$ 2.7} & \textbf{45.9 $\pm$ 1.0} & \multirow{3}{*}{44.3} \\  
                &   &  2  &  49.3 $\pm$ 3.1    & 44.4 $\pm$ 1.3 & 47.0 $\pm$ 1.0 & \underline{50.0 $\pm$ 0.8}  & \textbf{52.8 $\pm$ 0.9} &    \\  
                &   &  5  &  58.2 $\pm$ 2.5   & 57.0 $\pm$ 1.7 & 56.6 $\pm$ 0.7 & \underline{59.7 $\pm$ 1.9}  & \textbf{61.4 $\pm$ 0.4} &   \\ \cline{2-9}
                
                & \multirow{3}{*}{Flowers102}    & 1  & \underline{61.0 $\pm$ 3.7}     & 55.3 $\pm$ 4.2 & - & 56.7 $\pm$ 4.3 & \textbf{64.6 $\pm$ 5.2} & \multirow{3}{*}{67.3} \\  
                &   &  2  &  70.6 $\pm$ 1.4    & 66.8 $\pm$ 3.8 & 65.9 $\pm$ 1.4 & \underline{70.8 $\pm$ 2.1}  & \textbf{75.9 $\pm$ 1.3} &    \\  
                &   &  5  &  80.5 $\pm$ 1.5   & 81.6 $\pm$ 1.2 & 79.6 $\pm$ 2.1 & \underline{84.8 $\pm$ 0.7}  & \textbf{85.7 $\pm$ 0.9} &   \\ \cline{2-9}
                
                & \multirow{3}{*}{UCF101}    & 1  & 58.8 $\pm$ 2.7     & 60.0 $\pm$ 4.8 & - & \underline{61.3 $\pm$ 3.8} & \textbf{63.7 $\pm$ 1.2} & \multirow{3}{*}{64.3} \\  
                &   &  2  &  66.5 $\pm$ 0.8    & 67.4 $\pm$ 1.5 & 67.6 $\pm$ 1.9 & \underline{68.6 $\pm$ 0.7}  & \textbf{70.8 $\pm$ 0.9} &    \\  
                &   &  5  &  75.0 $\pm$ 0.7   & 74.0 $\pm$ 1.5 & 75.2 $\pm$ 1.2 & \underline{77.9 $\pm$ 0.4}  & \textbf{78.0 $\pm$ 0.9} &   \\ \cline{2-9}

                & \multirow{3}{*}{\textit{Average}}    & 1  &  \underline{52.9}   & 51.5 & - & 53.5 & \textbf{58.1} (\textcolor{blue}{+4.6}) & \multirow{3}{*}{58.6} \\  
                &   &  2  &  62.1   & 59.5 & 60.2 & \underline{63.1}  & \textbf{66.5} (\textcolor{blue}{+3.4}) &    \\  
                &   &  5  &  71.2   & 70.9 & 70.5 & \underline{74.1} & \textbf{75.0} (\textcolor{blue}{+0.9}) &    \\ \hline

\multirow{12}{*}{\centering\rotatebox{90}{MaPLe}} & \multirow{3}{*}{Textures}  & 1  & 37.1 $\pm$ 3.5     & 36.0 $\pm$ 7.8 & - & \underline{38.3 $\pm$ 0.7} & \textbf{45.8 $\pm$ 2.2} & \multirow{3}{*}{44.3} \\  
                &   &  2  &  \underline{46.0 $\pm$ 1.8}    & 37.8 $\pm$ 6.5 & 40.7 $\pm$ 0.4 & 45.4 $\pm$ 1.7  & \textbf{50.8 $\pm$ 2.0} &    \\  
                &   &  5  &  55.7 $\pm$ 2.0   & 54.9 $\pm$ 2.2 & 52.0 $\pm$ 1.2 & \textbf{57.7 $\pm$ 1.3}  & \underline{56.7 $\pm$ 0.9} &   \\ \cline{2-9}
                
                & \multirow{3}{*}{Flowers102}    & 1  & 62.2 $\pm$ 1.4     & 61.2 $\pm$ 2.5 & - & \textbf{67.0 $\pm$ 3.1} & \underline{66.0 $\pm$ 2.3} & \multirow{3}{*}{67.3} \\  
                &   &  2  &  \underline{72.0 $\pm$ 2.7}    & 64.0 $\pm$ 4.2 & 70.0 $\pm$ 4.4 & 70.1 $\pm$ 4.1  & \textbf{77.2 $\pm$ 1.9} &    \\  
                &   &  5  &  82.5 $\pm$ 2.5   & 83.0 $\pm$ 2.1 & 79.0 $\pm$ 2.6 & \underline{86.2 $\pm$ 0.2}  & \textbf{86.5 $\pm$ 0.4} &   \\ \cline{2-9}
                
                & \multirow{3}{*}{UCF101}    & 1  & \underline{64.4 $\pm$ 3.9}     & 59.7 $\pm$ 3.2 & - & 62.0 $\pm$ 0.7 & \textbf{65.8 $\pm$ 1.6} & \multirow{3}{*}{64.3} \\  
                &   &  2  &  69.6 $\pm$ 1.3    & 66.2 $\pm$ 3.1 & \underline{69.8 $\pm$ 0.1} & 68.8 $\pm$ 3.1  & \textbf{70.9 $\pm$ 2.4} &    \\  
                &   &  5  &  \underline{76.9 $\pm$ 0.8}   & 73.8 $\pm$ 2.4 & 76.3 $\pm$ 0.5 & 76.6 $\pm$ 0.8  & \textbf{77.6 $\pm$ 0.2} &   \\ \cline{2-9}

                & \multirow{3}{*}{\textit{Average}}    & 1  &  54.6   & 52.3 & -  & \underline{55.8} & \textbf{59.2} (\textcolor{blue}{+3.4}) & \multirow{3}{*}{58.6} \\  
                &   &  2  &  \underline{62.5}   & 56.0 & 60.2 & 61.4  & \textbf{66.3} (\textcolor{blue}{+3.8}) &    \\  
                &   &  5  &  71.7  & 70.6 & 69.1 & \underline{73.5} & \textbf{73.6} (\textcolor{blue}{+0.1}) &    \\
                \bottomrule
\end{tabular}}
$}
\end{table*}

\begin{table*}[t!]
\vskip-10pt\caption{\textbf{Results for ViT-L/14 and ResNet-50 Architectures.}}
\label{tab:arch_results}
\renewcommand{\arraystretch}{1}
    \centering
    \resizebox{0.7\textwidth}{!}{$
    \setlength{\tabcolsep}{1.2mm}{
\begin{tabular}{c|c|c|cccc|>{\columncolor{gray!20}}c|c}
\toprule
\textbf{Model} & \textbf{Dataset} & $B (\%)$ &  \textbf{Random} & \textbf{Entropy} \cite{wang2014new} & \textbf{CoreSet} \cite{sener2017active} & \textbf{BADGE} \cite{ash2019deep} &  \textbf{Ours} & \textbf{Zero-shot} \\ \hline

\multirow{12}{*}{\centering\rotatebox{90}{ViT-L/14}} & \multirow{3}{*}{Textures}  & 1  & \underline{45.7 $\pm$ 0.9}     & 40.0 $\pm$ 2.8 & - & 45.1 $\pm$ 2.5 & \textbf{55.1 $\pm$ 2.6} & \multirow{3}{*}{53.0} \\  
                &   &  2  &  50.0 $\pm$ 1.9    & 43.8 $\pm$ 2.0 & 51.8 $\pm$ 1.9 & \underline{52.1 $\pm$ 1.1}  & \textbf{56.4 $\pm$ 0.8} &    \\  
                &   &  5  &  60.4 $\pm$ 0.7   & 56.9 $\pm$ 2.0 & 59.7 $\pm$ 1.9 & \underline{61.7 $\pm$ 2.1}  & \textbf{63.7 $\pm$ 1.7} &   \\ \cline{2-9}
                
                & \multirow{3}{*}{Flowers102}    & 1  & 75.0 $\pm$ 1.3     & \underline{76.2 $\pm$ 0.9} & - & 72.6 $\pm$ 3.8 & \textbf{80.0 $\pm$ 1.4} & \multirow{3}{*}{79.3} \\  
                &   &  2  &  77.1 $\pm$ 1.7    & 73.2 $\pm$ 2.1 & 74.1 $\pm$ 2.1 & \underline{79.1 $\pm$ 4.0}  & \textbf{86.4 $\pm$ 2.0} &    \\  
                &   &  5  &  87.1 $\pm$ 2.4   & 87.8 $\pm$ 2.0 & 84.2 $\pm$ 2.2 & \underline{92.8 $\pm$ 0.2}  & \textbf{93.6 $\pm$ 0.7} &   \\ \cline{2-9}
                
                & \multirow{3}{*}{UCF101}    & 1  & 73.3 $\pm$ 1.6     & 72.4 $\pm$ 1.0 & - & \underline{73.7 $\pm$ 1.2} & \textbf{74.9 $\pm$ 1.3} & \multirow{3}{*}{74.2} \\  
                &   &  2  &  \underline{76.4 $\pm$ 0.7}    & 72.3 $\pm$ 1.4 & 74.1 $\pm$ 1.5 & 75.1 $\pm$ 2.6  & \textbf{77.5 $\pm$ 0.7} &    \\  
                &   &  5  &  \underline{82.4 $\pm$ 1.7}   & 80.0 $\pm$ 0.8 & 80.5 $\pm$ 0.3 & 82.0 $\pm$ 0.9  & \textbf{82.5 $\pm$ 1.5} &   \\ \cline{2-9}

                & \multirow{3}{*}{\textit{Average}}    & 1  &  \underline{64.7}   & 62.9 & - & 63.8 & \textbf{70.0} (\textcolor{blue}{+6.2}) & \multirow{3}{*}{68.8} \\  
                &   &  2  &  67.8   & 63.1 & 66.7 & \underline{68.8}  & \textbf{73.4} (\textcolor{blue}{+4.6}) &    \\  
                &   &  5  &  76.6   & 74.9 & 74.8 & \underline{78.8} & \textbf{79.9} (\textcolor{blue}{+1.1}) &    \\ \hline

\multirow{12}{*}{\centering\rotatebox{90}{ResNet-50}} & \multirow{3}{*}{Textures}  & 1  & 30.5 $\pm$ 2.6     & 26.8 $\pm$ 6.7 & - & \underline{33.2 $\pm$ 2.5} & \textbf{34.1 $\pm$ 0.5} & \multirow{3}{*}{40.4} \\  
                &   &  2  &  \underline{38.9 $\pm$ 1.1}   & 34.9 $\pm$ 1.7 & 38.9 $\pm$ 2.6 & 37.5 $\pm$ 1.1  & \textbf{43.0 $\pm$ 2.4} &    \\  
                &   &  5  &  48.0 $\pm$ 1.6   & 46.2 $\pm$ 1.2 & 44.9 $\pm$ 1.0 & \underline{48.4 $\pm$ 2.1} & \textbf{49.2 $\pm$ 2.3} &   \\ \cline{2-9}
                
                & \multirow{3}{*}{Flowers102}    & 1  & \underline{48.5 $\pm$ 3.8}     & 41.8 $\pm$ 3.8 & - & 45.6 $\pm$ 2.3 & \textbf{53.6 $\pm$ 2.0} & \multirow{3}{*}{62.1} \\  
                &   &  2  &  \underline{62.2$\pm$ 2.9}    & 55.7 $\pm$ 2.9 & 55.6 $\pm$ 2.1 & 60.1 $\pm$ 1.4  & \textbf{64.2 $\pm$ 2.9} &    \\  
                &   &  5  &  72.8 $\pm$ 2.3   & 71.4 $\pm$ 1.7 & 64.8 $\pm$ 0.4 & \underline{76.0 $\pm$ 1.4} & \textbf{77.7 $\pm$ 2.6} &   \\ \cline{2-9}
                
                & \multirow{3}{*}{UCF101}    & 1  & 41.8 $\pm$ 2.9     & 42.7 $\pm$ 2.5 & - & \underline{44.5 $\pm$ 0.6} & \textbf{51.0 $\pm$ 3.8} & \multirow{3}{*}{58.2} \\  
                &   &  2  &  55.6 $\pm$ 1.8    & 51.7 $\pm$ 1.2 & 54.9 $\pm$ 1.6 & \underline{56.2 $\pm$ 1.0}  & \textbf{58.3 $\pm$ 1.2} &    \\  
                &   &  5  &  66.2 $\pm$ 2.8   & 63.3 $\pm$ 2.3 & 65.2 $\pm$ 1.6 & \underline{66.5 $\pm$ 0.9} & \textbf{67.0 $\pm$ 1.0} &   \\ \cline{2-9}

                & \multirow{3}{*}{\textit{Average}}    & 1  &  40.3   & 37.1 & -  & \underline{41.1} & \textbf{46.2} (\textcolor{blue}{+5.1}) & \multirow{3}{*}{53.6} \\  
                &   &  2  &  \underline{52.2}   & 47.4 & 49.8 & 51.3  & \textbf{55.2} (\textcolor{blue}{+3.0}) &    \\  
                &   &  5  &  62.3  & 60.3 & 58.3 & \underline{63.6} & \textbf{64.6} (\textcolor{blue}{+1.0}) &    \\
                \bottomrule
\end{tabular}}
$}
\end{table*}

\begin{figure*}[t!]
    \begin{center}
        \includegraphics[width=0.85\linewidth, trim={3cm 1cm 2cm 0cm}]{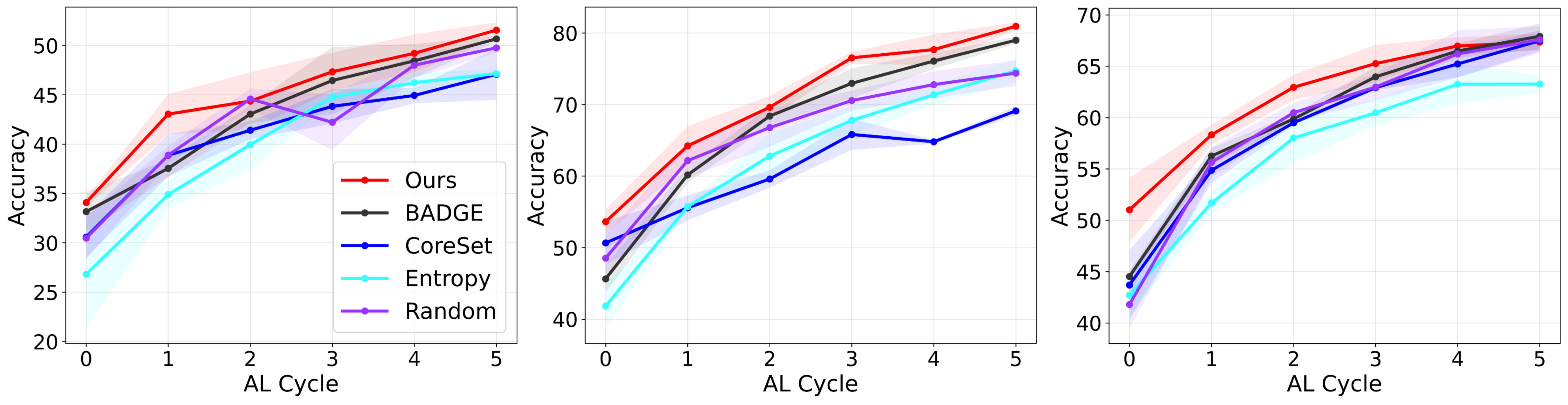}
    \end{center}
    \vskip -12.0pt
    \caption{ResNet-50 accuracy results over 6 AL cycles. From left to right: Textures, Flowers102, and UCF101 datasets.
    }
    \label{fig:rn50_curve} 
    \vskip -10.0pt
\end{figure*}

\noindent \textbf{Results for other architectures.} Previous results are achieved using ViT-B/16 architecture as the CLIP image encoder. Here, we conduct experiments utilizing ResNet-50 and ViT-L/14 as the CLIP image encoder, with the prompt tuning method set to CoOp. Table \ref{tab:arch_results} and Fig. \ref{fig:rn50_curve} present the classification results for this setting. These results show the superiority of our approach compared to other AL methods and verify the effectiveness of our method regardless of the backbone architecture used.

\subsection{Ablation Study}
\label{sec:ablations}

In this section, we conduct the ablation study on the Textures dataset to demonstrate the effectiveness of each component within our framework.\\
\noindent \textbf{Effect of different components.} In Fig. \ref{fig:ablation}, we conduct an extensive ablation study on the Textures dataset to verify the effectiveness of each component within our framework. In Fig. \ref{fig:ablation} (left), we evaluate the effect of our \textit{Entropy Calibration} and \textit{Neighbor Uncertainty} components. We can observe that eliminating either of them leads to a performance drop, which highlights the importance of these elements. In Fig. \ref{fig:ablation} (right), we study the impact of our proposed \textit{Entropy-Weighted Clustering} method. K-Means selects the nearest sample to the center from each cluster. Greedy Clustering selects the samples with the highest entropy from each cluster. We observe that our proposed entropy-weighted clustering method performs significantly better than such methods. This is because our method reduces the possibility of outlier selection. Even if an outlier shows high uncertainty, the increased number of inlier samples with relatively lower uncertainty will bias the cluster towards the high-density region of feature space, reducing the selection of outliers.\\
\noindent \textbf{Hyper-parameter analysis.} Our method introduces three hyperparameters: $N$ (in entropy calibration), $k$ (number of nearest neighbors), and $\alpha$ (in neighbor uncertainty). In Fig. \ref{fig:hyper_param}, we study the effect of these parameters for three different sampling ratios on the Textures dataset. From the left figure, we can observe that our entropy calibration benefits performance compared to the no calibration setting. From the middle and right figures, we see that our approach is robust to $k$ and $\alpha$, especially in higher sampling ratios. In our experiments, we fix the same value of $N$ for all datasets. For other hyper-parameters, we perform a grid search to determine the best values for each dataset. Specifically, we select $k$ from $\left \{10,20,50  \right \}$ and $\alpha$ from $\left \{0.05, 0.1,1  \right \}$. Finally, we run experiments with three different seeds and report the average results.

\noindent \textbf{Discussion on worse than zero-shot performance in certain cases.} From the results we can observe that the performance of AL methods, including ours, can be lower than zero-shot CLIP at $B=1\%$ due to limited labeled samples and overfitting from training the prompts for 50-200 epochs. However, starting from the second round, our approach significantly surpasses zero-shot performance in almost all datasets, with an average gain of $10\%$ at $B=2\%$ and $16\%$ at $B=5\%$ in CoOp experiments.

\section{Conclusion and Future Work}
In this paper, we propose a novel framework for addressing the problem of AL for prompt tuning of VLMs. Specifically, our approach includes an entropy calibration component that reduces the prediction bias towards frequently observed categories. We further combine the self-uncertainty of samples with their neighbor's uncertainty to provide a reliable uncertainty score. We leverage the uncertainty score to perform weighted clustering on the unlabeled data, leading to a diverse sampling of our approach. Moreover, we conduct extensive experiments on a wide range of specialized datasets where we significantly outperform zero-shot CLIP by only sampling a small number of unlabeled data for annotation. Our paper shows the importance of active learning in adapting VLMs to new datasets and suggests a promising solution for prompt tuning of such models. In future work, we plan to investigate the integration of our AL framework with other fine-tuning approaches.

\section* {Acknowledgements}
Research was sponsored by the Army Research Laboratory and was accomplished under Cooperative Agreement Number W911NF-23-2-0008. The views and conclusions contained in this document are those of the authors and should not be interpreted as representing the official policies, either expressed or implied, of the Army Research Laboratory or the U.S. Government. The U.S. Government is authorized to reproduce and distribute reprints for Government purposes notwithstanding any copyright notation herein.

{\small
\bibliographystyle{ieee_fullname}
\bibliography{egbib}
}

\end{document}